\def\year{2022}\relax
\documentclass[letterpaper]{article} 
\usepackage{aaai22}  
\usepackage{times}  
\usepackage{helvet}  
\usepackage{courier}  
\usepackage[hyphens]{url}  
\usepackage{graphicx} 
\usepackage{natbib}  
\usepackage{caption} 
\DeclareCaptionStyle{ruled}{labelfont=normalfont,labelsep=colon,strut=off} 
\usepackage{algorithm}
\usepackage{algorithmic}

\usepackage{newfloat}
\usepackage{listings}
\floatstyle{ruled}
\newfloat{listing}{tb}{lst}{}
\floatname{listing}{Listing}

\usepackage{subfig}
\usepackage{multirow}
\usepackage[raggedrightboxes]{ragged2e}

\usepackage{blindtext}
\usepackage{tcolorbox}
\usepackage{graphicx}
\usepackage{amsmath,physics}
\usepackage{hyphenat}

\newcommand{\revised}[1]{\textcolor{black}{#1}}

\usepackage[strict]{changepage}
\usepackage{framed}
\definecolor{background}{rgb}{0.96,0.96,0.96}
\newenvironment{nicequote}{%
  \MakeFramed{\advance\hsize-\width\FrameRestore}%
  \noindent\hspace{-4.55pt}
  \begin{adjustwidth}{}{7pt}%
  \vspace{2pt}\vspace{2pt}%
}
{%
  \vspace{2pt}\end{adjustwidth}\endMakeFramed%
}
\title{SocialQuotes: Learning Contextual Roles of Social Media Quotes on the Web}
\author{
    John Palowitch, Hamidreza Alvari, Mehran Kazemi, Tanvir Amin, Filip Radlinski
}
\affiliations{
    Google Research \\
    \{palowitch\footnote{Corresponding author},hamidrz,mehrankazemi,tanviramin,filiprad\}@google.com
}

\usepackage{bibentry}

\usepackage{xcolor}

\newcommand{\llm}{PaLM2}

\begin{document}

\maketitle

\begin{abstract}

\revised{Web authors frequently embed social media to support and enrich their content, creating the potential to derive \emph{web-based}, cross-platform social media representations that can enable more effective social media retrieval systems and richer scientific analyses. As step toward such capabilities, we introduce a novel language modeling framework that enables automatic annotation of \emph{roles} that social media entities play in their embedded web context. Using related communication theory, we liken social media embeddings to \emph{quotes}, formalize the page context as structured natural language signals, and identify a taxonomy of roles for quotes within the page context. We release SocialQuotes, a new data set built from the Common Crawl of over 32 million social quotes, 8.3k of them with crowdsourced quote annotations. Using SocialQuotes and the accompanying annotations, we provide a role classification case study, showing reasonable performance with modern-day LLMs, and exposing explainable aspects of our framework via page content ablations. We also classify a large batch of un-annotated quotes, revealing interesting cross-domain, cross-platform role distributions on the web.}

\end{abstract}

\section{Introduction}
\begin{figure}[ht]
\caption{\label{fig:schematic} Fictional web article with social media quotes from fictional platforms. We model social media quotation using a 4-stage procedure from communications theory~\citep{haapanen2020modelling}. The ``Societal localization'' stage is an unobserved process wherein the web author decides which \emph{roles} to seek for contextual support. We introduce a framework for inferring roles from the web context.}
\centering
\includegraphics[scale=0.30]{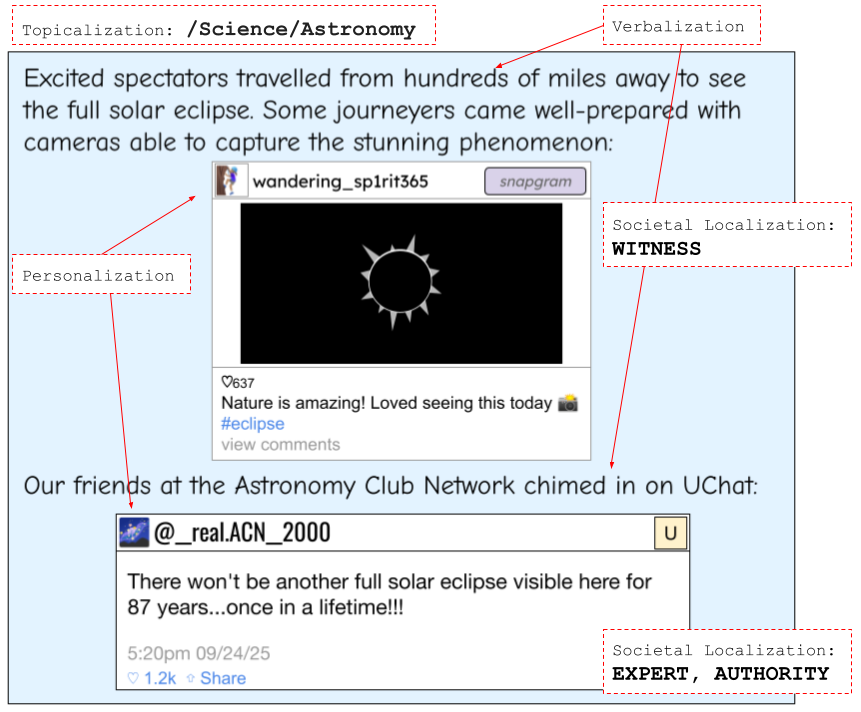}
\end{figure}

The global adoption of online social media (SM) \revised{platforms} over the past two decades has made \revised{them into \emph{de facto}} information stores of world knowledge\revised{, especially as primary sources of human stories, discussion, news commentary, and public announcements \citep{myers2014social}. As a result, web authors have increasingly cited SM in online articles and on the web at-large \citep{gearhart2014social}. Today, it is commonplace to find SM across the web in news stories, blog pieces, site info pages, and personal websites. There are standardized website tools for ``embedding'' SM in a page such that the post appears in-line, with platform-specific links and icons (see Figure \ref{fig:schematic}). Web authors may embed SM to exemplify, explain, provide evidence, or promote, potentially using ML-backed tools for SM source-seeking \citep{fernandes2023data}. Regardless, it is the web author that ultimately chooses the source appropriate for their page and frames the embedding with explanatory prose.}

\revised{We observe that, due to the ubiquity of social media embeddings, the web contexts of such embedded SM are potential sources of raw signals for learning representations of, and annotations on, SM posts and entities. Learning such representations and annotations could be valuable for (e.g.) building SM retrieval databases, or for feeding into scientific studies of society and the web. While many SM learning paradigms exist for \emph{on-platform} contexts \citep{balaji2021machine}, how to effectively learn from the \emph{web} context around SM embeddings remains an open challenge.}

\revised{As a step toward addressing this challenge, we introduce a natural-language learning framework for web-embedded social media, targeting automatic understanding of the ``role'' of the SM within the page context.}
As Figure~\ref{fig:schematic} illustrates, SM embeddings function as \emph{quotes} in that they can serve to tell a story, make a point, provide a reference, or show relevant human experiences from around the world, in some cases providing unique value beyond standard ``person-on-the-street'' sources \citep{gearhart2014social}. \revised{However, such ``roles'' cannot be immediately known from the web context -- they exist as latent, guiding categories in the mind of the web author.}
\revised{Our main hypothesis is that the web author's framing of SM quotes, as well as other web context around the page, can be used by a language model to infer the categorical role of the quote. If correct, modern-day language models can be profitably used to enhance SM databases with annotations about \emph{how} particular SM entities are commonly quoted across web topics. Our contributions are:}
\begin{enumerate}
    \item Using relevant communications theory, we link the web context of social media quotes to the social media entities involved, and propose a taxonomy of roles that SM quotes play in web contexts.
    \item We \revised{build and} release SocialQuotes\footnote{\url{https://www.kaggle.com/datasets/googleai/social-quotes}}, a data set computed from the Common Crawl with over 32M social media quotes, \revised{accompanied by} role labels for 8.3k quotes obtained from human annotators.
    \item \revised{We validate our main hypothesis with SocialQuotes by showing that an LLM can predict the} role of social quotes from their web context \revised{(and that performance improves with advanced reasoning techniques), and we provide a cross-domain, cross-platform case study analysis of role distributions across the web.}
\end{enumerate}

\revised{Two valuable aspects of our framework is that (1) it extends to all SM platforms commonly embedded on the web, and (2) it opens the door to modeling SM \emph{without any SM platform data}, drastically cheapening the cost of data collection for researchers pursuing certain applications.} We \revised{elaborate on these aspects and other} future directions in our closing section.

\section{Related Work}\label{sec:related}
\revised{To the best of our knowledge, our proposed framework is the first to rigorously establish a natural language learning task for SM embedded in the web. We identified five distinct sub-fields that inform our work in various ways. We give a brief overview of each field to correctly frame our work's novelty and impact.}

\subsubsection{Quotation in writing.} There is a large body of work in communications and linguistics on the \emph{form} and \emph{function} of traditional quoting in writing and media \citep{zelizer1995text, cope2020quoting, harry2014journalistic, bublitz2015introducing, zelizer1989saying, haapanen2020modelling}. Our work hearkens to this space by focusing on quotations of social media in the open web. Specifically, our proposed taxonomy of SM quotes aims to categorize the \emph{function} or 
``role'' of the quote rather than its \emph{form}, i.e.\ the specific language being used in the web context (though our data set release opens up this area as future work). As we describe in \revised{following sections}, our \revised{approach} builds on \citet{haapanen2020modelling}, \revised{allowing us to identify and use} a novel division of SM quote instances into \emph{societal groups} that \revised{web authors seek} from SM sources.

\subsubsection{Digital journalism.} Our work is also connected to the digital journalism space, in particular the study of \revised{SM posts appearing in online news media}. \citet{dumitrescu2021embedding}, \citet{gearhart2014social}, and \citet{broersma2014social} consider the differential \emph{effect} of SM quotes from certain SM platforms and accounts across media types and audiences. \citet{rony2018large}, \citet{kapidzic2022news}, and \citet{gruppi2021tweeting} study the differential \emph{frequency} with which SM sources are used in 
``unreliable''/tabloid-style outlets versus ``reliable''/mainstream outlets. \citet{mujib2022tweets} studies the \emph{speed} with which SM content is embedded in media outlets due to the ``pressures of the 24/7 news cycle''.

\revised{Many of these studies categorize social media accounts into ``types'', which correspond to what we call ``roles'' in this paper. Specifically,} among these studies, there are at least three \revised{distinct} 4-way taxonomies over the quoted social media accounts \cite{broersma2014social,kapidzic2022news,mujib2022tweets}. In each case, the authors manually classified SM accounts into the chosen taxonomy, and used the resulting codings in their scientific analyses. \revised{Our proposed learning paradigm aims to directly impact this field by enabling automatic classification of SM entities based on their embedded web context. As such, in the next section, we adopt a role taxonomy that both covers and extends each existing taxonomy in this space.}

\subsubsection{Quotation datasets.} There are several released web-based quotation data sets. \citet{tekir2023quote} propose a corpus of book quotes extracted from book reviews for the NLP task of automatic quote detection. \citet{vaucher2021quotebank} derive a large-scale corpus of traditional (non-SM) quotes from online news sources between 2008 and 2020. \revised{In the social media domain, embedded SM have not yet been likened to quotes, yet relevant datasets still exist.} \citet{mujib2020newstweet} release NewsTweet, a data set of embedded social media found in Google News sites. Our SocialQuotes data set expands upon NewsTweet in three ways: (1) \underline{ source}: \revised{by} using the Common Crawl, we \revised{are able to} cover SM quotes from a \revised{large random crawl of the web} rather than just news articles; (2) \underline{ scale}: our dataset covers approximately
\revised{12.7} million URLs with \revised{32.6} million embeddings, as opposed to approximately 69K articles and 136K embeddings from NewsTweet, and three platforms instead of one (NewsTweet covers Twitter only); (3)~\underline{ annotations}: we release topic metadata for all quotes and \revised{crowdsourced} role labels for a \revised{8.3k} subset of quotes.

\subsubsection{Influencer/expert detection.} \revised{The task of categorizing SM quotes into functional roles is related to the tasks} of influencer detection and expert finding\revised{; in fact, two roles in our chosen taxonomy are INFLUENCER and EXPERT}. Influencer detection, often achieved with graph learning algorithms \cite{zheng2020demand}, aims to recover structurally-central or abnormally-impactful nodes in social networks, either in general \cite{pei2020influencer} or with respect to given topics  \cite{panchendrarajan2023topic}. Expert finding is the task of retrieving members of a communication system (e.g.\ SM, Q\&A sites, email networks) who are knowledgeable or have skills in specialized areas, and is often performed with a mixture of NLP and graph-based methods \citep{balog2009language, balog2007determining, lin2017survey}. 
The main distinction between these paradigms and ours is that we do not attempt to classify SM users into certain roles \emph{per se}. Instead we attempt to classify the \emph{role} that a particular SM user's post plays in the context of a particular web page \revised{in which} their post is quoted.
A secondary \revised{and related} distinction is that we do not use any platform data: we \revised{study how} to use the \revised{surrounding web content} to infer the \revised{quote's} role.

\subsubsection{Social media in web data.} There have been some recent works that \revised{connet} social media entities and web data. For example, \citet{wen2023towards} develop the task of predicting WikiData \citep{vrandevcic2014wikidata} attributes of public figures using posts from the figures' social media accounts. Most related to our work, \citet{hombaiah2023tweembed} develop a retrieval model for Tweets\footnote{The previously-named Twitter platform is now called X, though in this publication we refer to X as ``Twitter'' and X posts as ``Tweets'' for consistency both with our source data and with past publications.} given a news article. They construct (but do not release) a data set of 8M (article, Tweet) tuples, each with at least one Tweet. They evaluate various two-tower retrieval models, some of which only process BERT encodings of the article and Tweet, and others which additionally include a Tweet account encoder taking in signals from the Twitter profiles. Our work both extends and complements this work. First, we release 37M instances of social media post embeddings in websites, which cover both news and non-news sites as well as additional platforms (TikTok and Instagram). One goal in releasing this dataset is to explicitly enable future research on retrieval models similar to those introduced in~\cite{hombaiah2023tweembed}, e.g.\ extending them to more platforms and more types of sites. Second, instead of retrieval, we study property prediction of the embedded social media entities using the surrounding page context. The next section formalizes our model of social media web embedding, which motivates our focus on the \emph{role} of the social post as the target of prediction.

\begin{table*}[t]
\caption{\label{tab:taxonomy} Social media quotation taxonomy. Each role is defined with respect to the web page author's intended function for the social media post within the context of the page they are writing, via the ``societal localization'' step.}
\begin{tabular}{|p{0.75in}|p{.9in}|p{2.6in}|p{2in}|}
\hline
Role Type & Role Name & Role Definition & Comparison Roles\\
\hline
\hline
\multirow{3}{*}{Elite} & EXPERT & Has professional experience, higher education, or valuable skills relevant to the page topic. & \emph{Expert} \cite{broersma2014social}, \emph{Journalist} \cite{mujib2022tweets}\\
\cline{2-4}
 & INFLUENCER & Is popular voice on the page topic and/or within the page's primary audience. & \emph{Cultural Producer} \cite{broersma2014social}, \emph{Celebrity} \cite{kapidzic2022news}, \emph{Personality}~\cite{mujib2022tweets}\\ 
\cline{2-4}
 & AUTHORITY & Is associated with a recognized societal position relevant to the page topic.  & \emph{Politician} \cite{broersma2014social}, \emph{Public Actor}, \emph{Media} \cite{kapidzic2022news}, \emph{Organization}, \emph{Media Outlet} \cite{mujib2022tweets}\\
 \hline
\multirow{3}{*}{\parbox{2cm}{Citizen}} & SUBJECT & The primary focus of the web page or article. & \emph{(none)}\\
\cline{2-4}
 & WITNESS & Was a witness or participant in an event described on the web page. & \emph{(none)}\\
\cline{2-4}
 & COMMENTER & Has shared thoughts or opinions about the page topic. & \emph{Vox Populi} \cite{broersma2014social}, \emph{Citizen} \cite{kapidzic2022news}\\
 \hline
\multirow{2}{*}{Commercial} & MARKETER & Is marketing products or selling services relevant to the page. & \emph{(none)}\\
\cline{2-4}
 & SELF-PROMOTER & Is the owner/author of the web page itself. & \emph{(none)}\\
 \hline
\end{tabular}
\end{table*}

\section{A Model of Social Quotation}\label{sec:roles}

In this paper, we draw an analogy between social media embeddings in web pages and rhetorical quotation in print, and use this to motivate a new \revised{NLP} task targeting the ``role'' of social media quotes. We base our approach on recent work in the communications literature which models quotation in news writing as a four-stage process \cite{haapanen2020modelling}: (1) ``topicalization'': the topic of the piece is established; (2) ``societal localization'': societal groups representing ``various roles (e.g.\ \emph{people concerned}, \emph{authorities})'' relevant to the topic are identified; (3) ``personalization'': representatives from the identified societal groups are chosen for inclusion in the piece; (4) ``verbalization'': the recorded views of the representatives are directly or indirectly woven into the piece according to the author's style and goals.

We adapt these four steps into a model of social media quotations on the web, such as those illustrated in Figure~\ref{fig:schematic}. Given a quotation instance $i$, we assume the author has arrived at a topic $t_i\in \mathcal{T} = \{T_1,\ldots,T_{n_t}\}$ (topicalization) for which they seek representative posts from social media. The author then chooses a set of roles $\rho_i = \{r_1, \ldots,r_k\}\subset \mathcal{R} = \{R_1,\ldots,R_{n_r}\}$ that indicates the 
``types'' of social media accounts being sought (societal localization) for worthwhile quotations on $t_i$. Next, the author chooses a social media post $p_i$ from a corpus $\mathcal{P} = \{P_1,\ldots,P_{n_p}\}$ created by a user $u_i$ that the author has determined to assume one or more roles from $\rho_i$ (personalization). Finally, the author rhetorically frames $p_i$ within the context of the web page, producing a novel text snippet $x_i$. 

\subsubsection{\revised{Role taxonomy.}}
In the above model, $\mathcal{P}$ can be any social media corpus, and $\mathcal{T}$ can be any set of web topic categories such the Cloud NL Categories\footnote{\url{https://cloud.google.com/natural-language/docs/categories}}. However, there is no standard set of social roles $\mathcal{R}$: indeed, our chosen modeling approach introduces a new focus on this aspect of quotation. Therefore a \revised{core aspect of} our work is \revised{to formulate an appropriate $\mathcal{R}$ for social media quotes on the web. To this end, we follow two complementary desiderata:}
\begin{enumerate}
    \item \revised{$\mathcal{R}$ should cover existing social account ``types'' adopted by digital journalism work (see ``Related Work''), so that our framework can be used to build models for similar efforts.}
    \item \revised{$\mathcal{R}$ should extend to SM embeddings that could plausibly be found outside of news or journalism pieces.}
\end{enumerate}
\revised{To achieve desiderata 1, we considered all social account ``types'' from \citet{kapidzic2022news}, \citet{broersma2014social}, and \citet{mujib2022tweets}, and found that they could be well-organized into four roles: EXPERT, INFLUENCER, AUTHORITY, and COMMENTER. A matching of the account types from past work to these roles is shown in Table \ref{tab:taxonomy}.}

\revised{To achieve desiderata 2, we propose to expand this initial role set in two ways, inspired by our manual review of web pages found in the Common Crawl corpus. First, we add SUBJECT and WITNESS roles to cover the many news and blog articles that (respectively) (1) focus on a certain entity and include SM quotes from that entity's account, or (2) focus on an event and include SM quotes from entities who participated or observed. Second, we add SELF-PROMOTER and MARKETER roles to cover (respectively) websites with (1) SM quotes from the same entity that owns the site (usually for promotional purposes), and (2) SM quotes that seem to be marketing a product or service.}

\revised{In total, our taxonomy has eight roles, as shown in Table~\ref{tab:taxonomy}. We observe that they can be divided into} Elite-type, Citizen-type, and Commercial-type roles, which we include in Table~\ref{tab:taxonomy} \revised{to illustrate the high-level semantic coverage of the taxonomy. In the next section, we empirically validate the taxonomy by providing an ``Other'' option to annotators. Over a random sample of SM quotes from the web, we found that annotators chose every role a non-trivial amount of times, and the ``Other'' option accounted for only $sim$1.4\% of all individual annotations.}

By adapting a fundamental model of traditional quoting to social media quotes on the web, we formalize the idea that the surrounding text of a social media quote can carry nontrivial information about the social media user via the ``societal localization'' step taken by the web author. Therefore, our primary methodological hypothesis (which we examine in our Experiments and Analysis section) is that the page context surrounding a SM quote can be used as signals to model the involved SM entities. In the next section, we describe the construction of the SocialQuotes data set, which we use in our experiments and release in full to the community.

\section{SocialQuotes Dataset}\label{sec:embeddings}

As illustrated in Figure~\ref{fig:schematic}, online content frequently embeds social media content. We define an SM embedding as the rendering of a social media post directly within the web page: it must be visible to a reader with a standard web browser without being logged in to any social media services (modulo any privacy restrictions that may be selectively enabled on various websites). An embedding is considered complete if it is possible to identify a canonical URL for the social media post as well as the author of the post. Motivated by our conceptual framework outlined in the previous section, we refer to social media embeddings as ``quotes'' throughout the remainder of the paper.

In this section we describe the construction of the SocialQuotes data set. Starting with the widely used Common Crawl corpus\footnote{https://www.commoncrawl.org/}, which consists of a frequently updated public multi-terabyte web crawl, we built a Python-based Apache Beam pipeline to identify and \revised{extract} SM quotes, and used a public language model to classify the surrounding web content into one or more topics. A sample of SM quotes were also annotated for social roles. We release\footnote{\url{https://www.kaggle.com/datasets/googleai/social-quotes}} the corpus of SM quotes and their annotations as the SocialQuotes dataset, as described fully in Appendix A Table~\ref{tab:schema}. For the purposes of this paper and dataset release, we focus \revised{on} Twitter, TikTok, and Instagram quotes, although our methodology is in no way limited to these platforms.

\subsection{Embedding Extraction Pipeline}\label{ss:pipeline}

We determined the HTML source patterns that create SM quotes from our chosen platforms through a trial-and-error process, inspecting web pages with visible SM quotes. The source patterns that we determined are shown in Appendix A Table~\ref{tab:embedding-patterns}. Given the crawled HTML from a webpage, the contained social media quotes that follow patterns that we identified can be extracted using BeautifulSoup\footnote{\url{https://www.crummy.com/software/BeautifulSoup/bs4/doc}} functions:
\begin{verbatim}
BeautifulSoup(
    html_doc, "html.parser")
.find_all(
    tag_type, class_=tag_class)
\end{verbatim}

Through this method, we extracted all instances of SM media quotes captured by these (\texttt{tag\_type}, \texttt{tag\_class}) pairs from the full Common Crawl 2022-05, 2022-40 and 2023-23 snapshots, restricting ourselves to web pages where a standard classifier indicated the content was primarily in English. Our extraction pipeline was written in Python using the Apache Beam framework, letting each worker handle SM embedding extraction from a single Common Crawl website record (identified with a unique URL). This pipeline yielded 32.6M embedded social media posts, with per-platform counts given in Table \ref{tab:embedding-patterns}.

\subsection{Context Parsing and Topic Annotation}\label{ss:context}

In addition to extracting social media quotes following patterns in Table~\ref{tab:embedding-patterns}, we also extracted the page context surrounding each quote by traversing the DOM tree in either direction until a pre-specified character limit $\tau$ was met. We extracted at least $\tau=300$ characters, and did not break in the middle of a DOM element such as a paragraph tag (thus some context snippets had more than $\tau$ characters). Note that we do not release context snippets in the SocialQuotes dataset, although these can be reconstructed from CommonCrawl and the SocialQuotes dataset. To enable the study of the intersection of SM entities and web topics, we classified each context snippet using the Google Cloud Content Classifier\footnote{\url{https://cloud.google.com/natural-language/docs/classifying-text}}. We release these topic annotations with SocialQuotes in the \texttt{context\_topics} field, and analyze their distribution with respect to SM quotes in our Experiments and Analysis section. To avoid releasing sensitive or inappropriate content, we filter the SocialQuotes entries according to rules described in Appendix A. \revised{This filtering brings our total quote count to 32,560,806, which is the size of our final SocialQuotes release.}

\subsection{Role Annotation Process}

From the collection of SM quotes identified, we sampled \revised{9k} SM quotes for annotation, choosing an equal number from each platform. To ensure that no URL was overly-represented (some URLs contain a huge number of SM quotes), we sampled quotes uniformly-at-random while ensuring that no URL was chosen twice. The quote instances were then each rated by 5 trained annotators as follows. 

First, the annotator was presented with the URL containing a citation extract, as well as the username of a social media author. We note that while SM quotes sampled were extracted from the \emph{static} Common Crawl collection, annotation was performed with annotators looking at \emph{live} web pages, which may differ from the crawled versions. Thus, we asked annotators to complete an initial task (full UI displayed in Appendix A, Figure~\ref{fig:cc-ui-q1}):

\begin{nicequote}
Please look at this webpage: \{{\tt Clickable URL}\}. Can you find a \{{\tt Platform}\} post by \{{\tt Username}\} in the main body of the page? This \{{\tt Clickable Post URL}\} by \{{\tt Username}\} may be embedded on the page.
\end{nicequote}

\revised{If the above question was answered in the negative, annotators were given a ``Not Found'' option to choose.} If the above question was answered affirmatively\footnote{Note the reference to the content being in the main body of the page: Annotators were instructed to \emph{not} annotate posts if they believed the post was embedded by someone other than the main page author, such as in a comment. Annotators were similarly trained to skip URLs with non-English content, or which did not load correctly.}, that the extracted citation can still be found, then a second question was shown asking the annotator to select from the social roles listed in Table~\ref{tab:taxonomy} as follows (full UI displayed in Appendix A, Figure~\ref{fig:cc-ui-q2}):

\begin{nicequote}
The web page author likely embedded the post(s) by the \{{\tt Platform}\} account \{{\tt Username}\} because they thought the account \_\_\_\_\_. Please choose {\bf all} options that apply. 
\end{nicequote}

For each option that was selected, the description from Table~\ref{tab:taxonomy} was presented directly in the annotation UI to remind the annotator of the exact meaning of each label. Further, annotators were asked to review a document providing examples of annotations for specific URLs prior to commencing the annotation process, and this document was made available to them to access at any time.

\begin{figure} 
    \centering
    \includegraphics[width=\columnwidth]{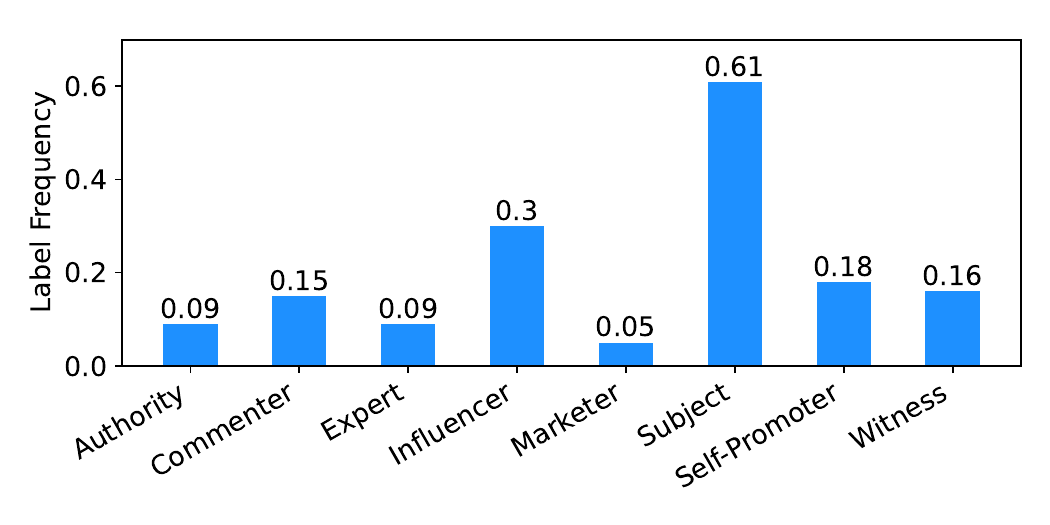}
    \caption{The frequency of each of the eight labels in the final dataset.}
    \label{fig:label_freq}
\end{figure}

\revised{We gave the annotators an ``Other'' option to indicate that they thought that the social post was playing a role potentially not covered by our taxonomy. In Appendix A, we provide additional statistics on the co-occurrence of the ``Other'' option with in-taxonomy roles (Table \ref{tab:fleiss-table}), as well as the frequency of the ``Not Found'' option (Figure \ref{fig:not-found-counts}). We find that the ``Other'' option was chosen only 289 times by any annotator, accounting for 1.4\% of annotator responses in which the post could be found. Furthermore, as shown by Figure \ref{fig:label_freq}, annotators chose all labels a non-trivial amount of times. These results validate the coverage of our chosen taxonomy.}

\textbf{Ethics statement.} We note that the human annotator work was carried out by participants who are paid contractors. Those contractors received a standard contracted wage, which complies with living wage laws in their country of employment. To ensure a consistent cultural understanding of social media, and to align with the fact that we only extracted social media from primarly English-language pages, the annotators were selected to be native English speakers based in the United States and in Canada.

\subsection{\revised{Ground-truth from annotations}}

\revised{We filter, aggregate, and validate annotations to produce ground-truth. First, any annotation on quotes proximal to sensitive content (see ``Context Parsing and Topic Annotation'' and Appendix A) are neither released nor used as ground-truth, leaving 8,281 annotations. From the remaining}, we define a ``valid'' annotation set for a given quote as one where at least two annotators were able to find the quote in the web page. \revised{Our annotation experiment produced 4,483 such sets.} To check annotator agreement, we report Fleiss' $\kappa$ \citep{fleiss1971measuring} across the valid annotations for each role in Table~\ref{tab:fleiss-table} in Appendix A. All roles had positive agreement, though with varying magnitudes. We release all completed annotations with at least one non-``Not Found'' result with the SocialQuotes data set under the field \texttt{role\_labels}, containing a JSON-formatted string encoding a dictionary over the roles in Table~\ref{tab:taxonomy}, giving the number of times annotators chose each role.

To derive ground-truth for our experiments in the next section, we add a role label $r$ to the ground-truth label set for a given example $i$ if at least two of the five annotators chose $r$ given example $i$. From the \revised{4,483} valid annotations, \revised{4,380} had at least one role in the ground-truth label set. We use these \revised{4,380} ground-truth annotations as test evaluation data for our experiments, which we cover in the next section. \revised{In Appendix A Figure \ref{fig:role-confusion}, we provide a confusion matrix over the roles showing how frequently roles co-occurred in a ground-truth quote.}

\section{Experiments and Analysis}\label{sec:experiments}
In this section we describe experiments that \revised{constitute a case-study on our main intended use-case for our framework: inferring and analyzing roles of social media entities in web contexts.} We consider two primary high-level research questions:
\begin{itemize}
    \item \textbf{RQ1}: Can the role of a social media post quoted in a web site be inferred from the surrounding web context?
    \item \textbf{RQ2}: Do roles, platforms, and users distribute across web domains and web topics in meaningful, intuitive, or surprising ways?
\end{itemize}
An affirmative answer to these questions would show that there is useful information about social media entities contained in the web content that quotes them, confirming our overall hypothesis laid out in the ``Model of Social Quotation'' section. These experiments also show the utility of SocialQuotes toward future research directions, as described further in our closing section.

\subsection{Role inference}\label{ss:inference}
Here we investigate \revised{our main hypothesis: whether} the role of a SM quote \revised{can be inferred by a language model} from the surrounding text and page context. \revised{We use a pre-trained LLM in our experiment as a representative state-of-the-art language model to expose the present-day capabilities on the role classification task. Comparisons between LLMs or between an LLM and smaller models are left for future work.} Specifically, we seek to answer the following questions:
\begin{itemize}
    \item \textbf{RQ1.1}: Can an LLM infer the role of \revised{SM quotes} from the surrounding text (rhetorical framing) and the page context?
    \item \textbf{RQ1.2}: Does \revised{using LLM reasoning techniques such as chain-of-thought \cite{wei2022chain} and self-consistency \cite{wang2022self} help with role classification?}
    \item \textbf{RQ1.3}: What elements of the page context are most important for \revised{role classification}?
\end{itemize}
As our LLM, we use Google's \llm\  \texttt{text-bison} model \citep{palm2}, which can be accessed via their Vertex AI\footnote{http://cloud/vertex-ai/docs/generative-ai/model-reference/text}. 

\subsubsection{Task Design.} As illustrated in Figure~\ref{fig:schematic}, in a web page with social media quotes, the web author can assign potentially multiple roles to any given social media quote. To handle this scenario, we design a task where \llm\ is asked to make a binary decision on each role individually. In other words, for a given social media quote, we prompt \llm\ eight times, one time for each role in our taxonomy, and then aggregate the answers into a single result. We now describe the design of the prompt and evaluation for this setting.

\subsubsection{Prompt Structure.} Our \llm\ prompt \revised{(shown in the Appendix)} includes \revised{(1)} a preamble, \revised{(2)} role-specific  elements from the page (including the text surrounding the quote, the URL of the page, the URL of the social media post, and the handle of the social media post), and optionally \revised{(3)} few-shot examples. The \texttt{<role\_specific\_binary\_prompt>} field is filled by a question that is specific to the current role being decided upon. The question contains a description of the role resembling those in Table~\ref{tab:taxonomy}. We list all eight questions in Table~\ref{tab:prompt-table}, Appendix B:

\revised{To construct the prompt,} the common crawl URL, social post URL, and extracted profile handle are each extracted from their corresponding SocialQuotes data set fields listed in Table~\ref{tab:schema}. Note that the extracted profile handle is a parse of social user profile, which is a URL to the profile page. The quote context field contains the quote context as described in our previous ``Context Parsing'' section. The few shot examples block contains five examples pulled from our SocialQuotes corpus, containing all four fields, and a final line reading 
``The answer is \{yes,no\}''. This final line instructs \llm\ how to give a binary answer to the prompt.

\subsubsection{Evaluation.} We evaluate the binary judgements given by \llm\ against our annotated examples using the standard F1 score over each of the eight roles in our taxonomy. As baselines, we compare against the following:
\begin{enumerate}
    \item True-frequency: a role $r$ is a predicted positive with probability $p_r$, where $p_r$ is the ground-truth probability of role $r$. Note that under this baseline, the F1 score converges in probability to $p_r$ given an asymptotic number of samples. Figure~\ref{fig:label_freq} shows the $p_r$ values for each $r$.
    \item Coin-flip: a role $r$ is a predicted positive with probability 0.5. Under this baseline, the F1 score converges to $p_r / (0.5 + p_r)$. We list this value in the ``Coin flip'' column in Table 3.
\end{enumerate}

\subsubsection{Performance analysis.} The results in the ``Fewshot'' column of Table 3 show the F1 scores of \llm\ provided only with the preamble, the fewshot examples, and the social media quote data fields. We find that \llm\ outperforms the coin-flip baseline on every role, sometimes with a great margin. This provides an affirmative answer to RQ1.1: there seems to be natural language signals present in the page context that can help infer the societal localization step of  quotation, which we described in our 
``Model of Social Quotation'' section. Of note, this also provides important validation for our approach that attempts to connect traditional quotation with the process of social media embeds on the web.

\subsubsection{Chain-of-thought prompting.} To investigate whether providing reasoning examples helps the model, we re-ran the experiment while adding a ``chain-of-thought'' (CoT) paragraph \cite{wei2022chain} immediately following each few-shot example. Each CoT paragraph used the given data fields to explain in common prose why the answer was yes or no. An example of a chain-of-thought paragraph is \revised{shown in the Appendix (see ``Prompt Structure'' section)}:

From Table 3, it is clear that CoT prompting \revised{improved} the classification results in every role, \revised{leading to increased} macro-average F1. \revised{Even so,} we notice that providing CoT leads to an over-prediction of the positive class leading to a higher recall and a lower precision compared to the Fewshot model \revised{(see Table 8 in the Appendix)}. We next examine some methodology to remedy this effect.

\subsubsection{Self-Consistency and Persistence.} Self-Consistency \cite{wang2022self} is a technique used for LLM-based classification where instead of calling the model once in a greedy mode and taking its output as the final answer, one calls the model multiple times and takes the majority vote of its predictions. The model is called with temperature\footnote{\revised{Larger values of the ``temperature'' parameter in calls to standard LLMs induce more randomness in the response, whereas zero temperature results in determinism.}} \revised{equal to 0.5} to enable generating non-identical but highly probable samples. We tested a variant of our Fewshot and CoT models with self-consistency where besides the greedy model output, we generated another $3$ sets of model samples at a temperature of $0.5$. Observing that our CoT model has a high recall but low precision, we tested a special case of self-consistency, which we call \textbf{``Persistence''}, where we set the prediction for each example to be ``yes'' only if all model calls predict ``yes''. The results are reported in Table 3. We observe that both self-consistency and persistence boost the performance of the CoT model, but they are not as effective for the Fewshot model. Specifically, CoT plus persistence achieves the best results in terms of macro-average F1 across all roles. \revised{This answers \textbf{RQ1.2} in the affirmative.}

\subsubsection{Ablation study.} Finally, \revised{to address \textbf{RQ1.3},} we examine which parts of the page context contribute most to the performance of \llm. At a conceptual level, the URL should provide web author information, the post URL should provide platform information, the handle should provide social media author information, and the snippet should provide role-level information about the social media post. We ablate each of these signals and re-run the experiment, with results in Table 8. Looking at the macro-average scores, we find that all signals contribute some performance value to the model, however the snippet provides the most. This accords with our fundamental hypothesis laid out in our 
``Model of Social Quotation'' section, which is that the actual language used to quote the social media author can provide key information about the role that the post is playing in the page.

\begin{table*}[t]
\caption{\label{tab:inference-results} F1-Scores for \llm\ on social roles classification with different prompt variants. We also report the expected F1 score under a random coin flip. The best results are highlighted in bold. \emph{Fewshot} corresponds to providing fewshots with no CoT, \emph{CoT} corresponds to providing fewshots with CoT, \emph{SC} corresponds to self-consistency, and \emph{P} corresponds to persistence.
}
\centering
\begin{tabular}{|c|c|c|c|c|c|c|c|c|}
\hline
Role $\downarrow$, Model $\rightarrow$ & Coin flip & Fewshot & Fewshot + SC & Fewshot + P & CoT & CoT + SC & CoT + P \\
\hline
\hline
AUTHORITY	 & 0.16 & 0.234 & 0.226 & 0.246 & 0.239 & 0.227 & \textbf{ 0.277 } \\ \hline
COMMENTER	 & 0.237 & 0.255 & 0.257 & 0.252 & 0.28 & 0.277 & \textbf{ 0.289 } \\ \hline
EXPERT	 & 0.155 & 0.179 & 0.178 & 0.189 & 0.19 & 0.186 & \textbf{ 0.205 } \\ \hline
INFLUENCER	 & 0.371 & 0.454 & 0.457 & 0.45 & \textbf{ 0.461 } & 0.458 & \textbf{ 0.461 } \\ \hline
MARKETER	 & 0.096 & 0.236 & 0.232 & 0.237 & 0.282 & 0.29 & \textbf{ 0.344 } \\ \hline
SUBJECT	 & 0.548 & 0.573 & 0.591 & 0.499 & 0.727 & \textbf{ 0.732 } & 0.71 \\ \hline
SELF-PROMOTER	 & 0.27 & 0.562 & 0.627 & 0.37 & 0.629 & 0.634 & \textbf{ 0.721 } \\ \hline
WITNESS	 & 0.246 & 0.286 & 0.287 & 0.282 & 0.302 & 0.301 & \textbf{ 0.307 } \\ \hline\hline
Macro Average	 & 0.26 & 0.347 & 0.357 & 0.316 & 0.389 & 0.388 & \textbf{ 0.414 } \\ \hline

\end{tabular}
\end{table*}

\begin{table*}[t]
\centering
\caption{\label{tab:ablation-results} Ablation results for social roles classification. Measuring performance by removing one component at a time.
}
\begin{tabular}{|c|c|c|c|c|c|}
\hline
Role $\downarrow$, Removed Field $\rightarrow$ & None & URL & HANDLE & POST URL & SNIPPET \\
\hline
\hline
AUTHORITY	 & 0.234 & 0.253 & 0.215 & 0.22 & 0.307 \\ \hline
COMMENTER	 & 0.255 & 0.294 & 0.233 & 0.242 & 0.255 \\ \hline
EXPERT	 & 0.179 & 0.188 & 0.184 & 0.177 & 0.164 \\ \hline
INFLUENCER	 & 0.454 & 0.455 & 0.46 & 0.454 & 0.348 \\ \hline
MARKETER	 & 0.236 & 0.253 & 0.187 & 0.263 & 0.283 \\ \hline
SUBJECT	 & 0.573 & 0.546 & 0.532 & 0.521 & 0.496 \\ \hline
SELF-PROMOTER	 & 0.562 & 0.296 & 0.555 & 0.566 & 0.626 \\ \hline
WITNESS	 & 0.286 & 0.292 & 0.279 & 0.269 & 0.145 \\ \hline\hline
Macro Average	 & 0.347 & 0.322 & 0.331 & 0.339 & 0.328 \\ \hline

\end{tabular}
\end{table*}

\subsection{Social quotes analysis}\label{ss:analysis}
In this section we use the SocialQuotes data set to provide novel, cross-platform, cross-role insights into the landscape of social quotation on the web. We first use \llm\ prompted with all page context signals and CoT to infer the roles of a large batch of social quotes (approx. 117k quotes). We use the full data set and these inferred-on quotes to investigate the following questions:
\begin{itemize}
    \item \textbf{RQ2.1}: Do websites covering certain \textbf{topics} tend to favor quoting certain social media platforms or roles?
    \item \textbf{RQ2.2}: Do websites at certain \textbf{domain names} tend to favor quoting certain social media platforms or roles?
\end{itemize}
We translate these questions into the following general measurement problem: given a website attribute $a$ and a social media attribute $x$, do websites with attribute $a$ (for example, a certain domain like news.yahoo.com) disproportionately quote social media posts with attribute $x$ (for example, a certain platform like Twitter)? We use two metrics to examine this question: the relative proportion of $x$ found with attribute $a$, and the mutual information between $x$ and $a$. Specifically, \revised{define $N$ as the total number of quotes in the corpus}, $N_a$ as the number of quotes in the corpus such that the website has attribute $a$, $N_x$ similarly for attribute $x$, and $N_{ax}$ as the number of quotes such that the website has attribute $a$ and the SM post has attribute $x$. Then we define $p(a, x) = \frac{N_{ax}}{N_a}$ as the relative proportion and $\text{MI}(a, x) = N\frac{N_{ax}}{N_aN_x}$ as the mutual information. The mutual information metric in particular represents how much more often sites with attribute $a$ quote social media with attribute $x$ compared with how often sites quote social media with attribute $x$ overall.

We investigate \textbf{RQ2.1} and \textbf{RQ2.2} by computing $p(a,x)$ and $\text{MI}(a,x)$ for $a=$ domains and topics and for $x=$ platforms, handles, and roles. In Tables~\ref{tab:domain-analysis-results} and \ref{tab:topic-analysis-results}, we report results for some of the most popular domains and topics in the SocialQuotes data set, hand-picked for diversity across a variety of web topics and news markets. We rank platforms and roles by $\text{MI}$ score, reporting all three platforms and only the top-3 roles. We note that for platforms, these statistics are computed across the entire SocialQuotes corpus, as no role annotations are needed. For roles, we use the 117k batch of inferred roles.

\subsubsection{Results.} We report topic-based results in Table~\ref{tab:topic-analysis-results} and domain-based results in Table~\ref{tab:domain-analysis-results} (Appendix B). Our main observation is the following: both the topics and the domains we list can be roughly split into \textbf{culture}-oriented topics/domains and \textbf{reporting}-oriented topics/domains, and these groups tend to favor quotes from certain platforms and certain roles. In particular:

\begin{itemize}
    \item Culture-related topics (e.g.\ /Beauty, /Shopping, /Food \& Drink, /Travel, /Celebrities) and domains (e.g.\ allure.com, upworthy.com) tend to quote  preferentially from Instagram and TikTok, and preferentially from Marketer, Influencer, and Self-Promoter roles. This follows the intuition that culture-related web pages should desire social media quotes from accounts that blog about new fashion, art, and culinary trends, which tend to occur more on Instagram/TikTok and assume commercial/Influencer-type roles.
    \item On the other hand, reporting-related topics (e.g.\ /News, /Health) and domains (e.g. sportingnews.com, cnn.com, finance.yahoo.com) tend to quote preferentially from Twitter and from Authority, Commenter, Expert, and Witness roles. This \revised{aligns with research} that Twitter often functions as a news site \revised{for many users} \citep{kwak2010twitter, porter2023twitter}, with the aforementioned roles micro-blogging about their areas of expertise/authority, or about their day-to-day experiences.
\end{itemize}

Our second observation is simpler, yet striking: almost all of these domains and topics quote from all three platforms. This supports one of our basic motivations, which is that social media quoting is now a cross-platform phenomenon. We release the SocialQuotes data set to the community so that insights such as the above (and beyond) can be studied further.

\begin{table*}[t]
\centering
\caption{\label{tab:topic-analysis-results} Table of co-occurrence statistics for topics vs.\ platforms and roles. Note that the top-3 roles and the (only) 3 platforms do not correspond to each other: roles and platforms are independently ranked by MI.
}
\begin{tabular}{|l||l|l|l||l|l|l|}
\hline
Topic ID & Platforms & $p$ & MI & Roles & $p$ & MI \\
\hline
\hline

/News/Sports News& Twitter & 1E+0 & 1.2 & Authority & 8E-1 & 5.3 \\
& Instagram & 4E-2 & 0.2 & Commenter & 9E-1 & 5.1 \\
& TikTok & 9E-4 & 0.1 & Expert & 8E-1 & 4.9 \\
\hline
\hline

/Arts \& Entertainment/Celebrities \& Entertainment News& Instagram & 4E-1 & 1.9 & Influencer & 8E-1 & 5.3 \\
& TikTok & 1E-2 & 1.0 & Witness & 2E-1 & 5.2 \\
& Twitter & 6E-1 & 0.8 & Commenter & 9E-1 & 5.1 \\
\hline
\hline

/Beauty \& Fitness/Fashion \& Style& Instagram & 8E-1 & 4.1 & Marketer & 4E-1 & 11.9 \\
& TikTok & 2E-2 & 1.5 & Influencer & 9E-1 & 5.9 \\
& Twitter & 2E-1 & 0.2 & Subject & 8E-1 & 5.3 \\
\hline
\hline

/News/Politics& Twitter & 1E+0 & 1.3 & Authority & 8E-1 & 5.5 \\
& TikTok & 1E-3 & 0.1 & Commenter & 9E-1 & 5.0 \\
& Instagram & 1E-2 & 0.1 & Expert & 9E-1 & 4.9 \\
\hline
\hline

/Shopping/Apparel& Instagram & 7E-1 & 3.4 & Marketer & 6E-1 & 17.9 \\
& TikTok & 3E-2 & 1.8 & Influencer & 9E-1 & 5.6 \\
& Twitter & 3E-1 & 0.4 & Self-Prom. & 5E-1 & 5.3 \\
\hline
\hline

/Travel \& Transportation/Tourist Destinations& Instagram & 7E-1 & 3.7 & Marketer & 3E-1 & 9.6 \\
& TikTok & 2E-2 & 1.6 & Witness & 3E-1 & 6.9 \\
& Twitter & 3E-1 & 0.3 & Self-Prom. & 5E-1 & 5.2 \\
\hline
\hline

/Food \& Drink/Cooking \& Recipes& TikTok & 6E-2 & 3.8 & Marketer & 6E-1 & 18.9 \\
& Instagram & 7E-1 & 3.7 & Self-Prom. & 6E-1 & 6.0 \\
& Twitter & 2E-1 & 0.3 & Influencer & 8E-1 & 5.3 \\
\hline
\hline

/Health/Public Health& Twitter & 1E+0 & 1.2 & Authority & 8E-1 & 5.7 \\
& TikTok & 3E-3 & 0.2 & Expert & 9E-1 & 5.0 \\
& Instagram & 3E-2 & 0.1 & Subject & 8E-1 & 5.0 \\
\hline
\hline

\end{tabular}
\end{table*}

\section{Discussion}\label{sec:discussion}
In this paper, we \revised{introduced a new NLP paradigm for classifying social media posts that appear in web contexts}. We \revised{drew an equivalence between the social media} embedding process \revised{and} quoting, using a formal model of quotation, \revised{and we identified} a latent taxonomy of \emph{roles} that social media posts play when they are quoted on the web. \revised{We built and released SocialQuotes, a dataset of $\sim$32M social media embeddings from $\sim$12M web pages}. With \revised{carefully-designed experiments on SocialQuotes}, we showed that roles can indeed be predicted from the web page context, and that the distribution of platforms, accounts, and roles over web domains can be profitably studied \revised{with our framework}. We now discuss limitations of our work, future directions, and potential use-cases for SocialQuotes.

\subsection{Limitations}\label{ss:limitations}
There are three overarching limitations of our work\revised{, which we discuss below}.

\subsubsection{Coverage and bias of SocialQuotes.} \revised{Our dataset is limited by our coverage of social platforms and the reach of Common Crawl, among other factors:}
\begin{enumerate}
    \item \revised{SocialQuotes} is derived only from three Common Crawl snapshots in the year range 2022-2023, out of hundreds of such snapshots spanning over a decade.
    \item SocialQuotes only covers English-language websites, and only three platforms (Twitter, TikTok, and Instagram).
    \item Our \revised{approach} likely does not cover all social media quote HTML signatures for \revised{the three platforms}.
    \item \revised{Common Crawl does not visit  web sites that were paywalled at crawl time, and likewise our annotators could not annotate sites that were paywalled at annotation time (nor can they rate SM posts that have become private since the crawl).}
\end{enumerate}
\revised{Due to the above caveats, our empirical results from SocialQuotes hold only for a subset of the web and a subset of SM quotes, in particular those parts of the web and SM that can be found without logins.} We plan to release an \revised{updated} version of SocialQuotes that expands on \revised{some of} these dimensions\revised{; however, other dimensions such as paywalled sites and other crawl restrictions are strict limitations. Fortunately, our framework is fully-transferable to any given web corpus, including web corpi that can be built through services other than Common Crawl that may be able to access proprietary websites.}

\subsubsection{Role taxonomy.} While we consider our proposed quote taxonomy to be a well-grounded \revised{facet of our framework, covering journalistic roles used by}  existing analyses of SM quotes \revised{in online news, while also extending to other roles}, it is by no means the only possible taxonomy, nor is it necessarily complete or useful for every given application. \revised{In particular, because our taxonomy was inspired partially by our manual review of Common Crawl websites, and Common Crawl has its own limitations (see above), it is possible we are not aware of certain roles, or better formulations of our taxonomy. Nonetheless, the we believe the taxonomy we chose was valid within our overall framework, and will be useful for practitioners and researchers studying web domain distributions similar to that found in Common Crawl.} Our release of SocialQuotes --  only a small percentage of which is labelled with our taxonomy -- will enable further research into the many possible ways that SM authors are cited by web authors.

\subsubsection{Source trustworthiness.} The third limitation of our data set and our overall approach is that some web authors may have minority opinions about appropriate roles for certain social media accounts or posts. In general, there is room for reasonable disagreement when characterizing a quote: whether that quote is a social media post quoted in a website, or a real quote in a news article. However, we see this as a microcosm of larger, unavoidable issues encountered when learning over large-scale web data: knowledge graph entities are rarely discussed consistently, and factuality remains a broad challenge \citep{augenstein2023factuality}. The SocialQuotes dataset enables the study of these issues in the social media realm.

\subsection{Long-term impact and use-cases}\label{ss:future-work}
\revised{We consider three primary potential impacts and use-cases of our framework.}

\subsubsection{Learning SM annotations for retrieval and analysis.} One primary impact of our work is that it unifies social media data from many platforms in a common annotation space, derived from the open web. As we showed in \revised{our case study}, our data set and role taxonomy allows researchers to study cross-platform patterns over the web, generating empirical results that cover novel intersections of social media populations. \revised{By adopting our framework and LLM prompt scheme, digital journalism researchers may automatically annotate large batches of SM on the web, rather than performing the hand-coding involved in the work we discussed earlier in this paper.} \revised{More broadly, using our paradigm,} developers may be able to improve tools that allow analysts, journalists, \revised{and researchers to retrieve clusters of social media entities -- from multiple platforms at once -- that pertain to a specific topic and functional role in web contexts.}

\subsubsection{Toward independence from SM APIs.} \revised{A strong benefit of our framework} is that \revised{it allows for} modeling social media with\revised{out depending on} social platform APIs. Recent restrictions on these APIs \citep{paresh2023reddit, porter2023twitter, lawler2022meta} have encouraged SM researchers to imagine a 
``post-API'' landscape \citep{perriam2020digital, tromble2021have, freelon2018computational}. Our work provides a foundation for modeling social media entities purely with publicly-available, off-platform data.

In some areas such as influencer detection, this also paves the way for more rigorous experimental design. In many influencer detection studies, influencer labels are derived from the same platform signals that feed into the model. For example, \citet{zheng2020demand} suggest that a Twitter account should receive a positive evaluation label for a particular topic if (1) greater than 50\% of their Tweets are related to the topic, and (2) their topic-specific Tweets are among the most-retweeted on that topic. However, the unsupervised influencer detection approach introduced in the paper uses both post-topic semantic similarity and the retweet graph as \revised{model inputs}. Similarly, \citet{mittal2020social} evaluate using platform signals that are explicitly used as features in the detection model. This evaluation strategy (additionally observed in similar studies such as \citet{kim2016topical} and \citet{oro2017detecting})  does not establish a task with meaningful headroom for models to achieve.
\revised{Our framework and dataset} provide a potential solution to this problem by establishing an \revised{off-platform} source of labels for influencer detection models. \revised{Of course, web authors who quote a particular SM may do so because the SM entity has many followers on the platform, in which case the platform signal would be confounding. However, even accounting for this, the follower count is just one potential signal that the web author could be influenced by, making their quote a non-trivial label for an on-platform influencer model (as opposed to a label derived purely from platform signals)}. Moreover, our role taxonomy opens the door for modeling many other types of social accounts beyond influencers.

\subsubsection{Toward generative citation.} The SocialQuotes data set also exposes pointers (via Common Crawl URLs) to examples of humans writing and reasoning about social media posts in relation to their web content. This can enable the study of the \emph{language} of web quoting and citation, and eventually lead to LLM systems that can weave relevant user-generated content into generated text with natural, purposeful, and well-reasoned exposition. We see this as a speculative yet strongly promising area for future work.

\bibliographystyle{ACM-Reference-Format}
\bibliography{main}

\begin{thebibliography}{43}
\providecommand{\natexlab}[1]{#1}

\bibitem[{Augenstein et~al.(2023)Augenstein, Baldwin, Cha, Chakraborty, Ciampaglia, Corney, DiResta, Ferrara, Hale, Halevy et~al.}]{augenstein2023factuality}
Augenstein, I.; Baldwin, T.; Cha, M.; Chakraborty, T.; Ciampaglia, G.~L.; Corney, D.; DiResta, R.; Ferrara, E.; Hale, S.; Halevy, A.; et~al. 2023.
\newblock Factuality challenges in the era of large language models.
\newblock \emph{arXiv preprint arXiv:2310.05189}.

\bibitem[{Balaji, Annavarapu, and Bablani(2021)}]{balaji2021machine}
Balaji, T.; Annavarapu, C. S.~R.; and Bablani, A. 2021.
\newblock Machine learning algorithms for social media analysis: A survey.
\newblock \emph{Computer Science Review}, 40.

\bibitem[{Balog, Azzopardi, and de~Rijke(2009)}]{balog2009language}
Balog, K.; Azzopardi, L.; and de~Rijke, M. 2009.
\newblock A language modeling framework for expert finding.
\newblock \emph{Information Processing \& Management}, 45(1): 1--19.

\bibitem[{Balog, De~Rijke et~al.(2007)}]{balog2007determining}
Balog, K.; De~Rijke, M.; et~al. 2007.
\newblock Determining Expert Profiles (With an Application to Expert Finding).
\newblock In \emph{IJCAI}, volume~7, 2657--2662.

\bibitem[{Broersma and Graham(2014)}]{broersma2014social}
Broersma, M.; and Graham, T. 2014.
\newblock Social media as beat: Tweets as a news source during the 2010 British and Dutch elections.
\newblock In \emph{Online Reporting of Elections}, 113--129. Routledge.

\bibitem[{Bublitz(2015)}]{bublitz2015introducing}
Bublitz, W. 2015.
\newblock Introducing quoting as a ubiquitous meta-communicative act.
\newblock In Arendholz, J.; Bublitz, W.; and Kirner-Ludwig, M., eds., \emph{The pragmatics of quoting now and then}, 1--26. De Gruyter Mouton Berlin.

\bibitem[{Cope(2020)}]{cope2020quoting}
Cope, J. 2020.
\newblock Quoting to persuade: A critical linguistic analysis of quoting in US, UK, and Australian newspaper opinion texts.
\newblock \emph{AILA Review}, 33(1): 136--156.

\bibitem[{Dumitrescu and Ross(2021)}]{dumitrescu2021embedding}
Dumitrescu, D.; and Ross, A.~R. 2021.
\newblock Embedding, quoting, or paraphrasing? Investigating the effects of political leaders’ tweets in online news articles: The case of Donald Trump.
\newblock \emph{New Media \& Society}, 23(8): 2279--2302.

\bibitem[{Fernandes, Moro, and Cortez(2023)}]{fernandes2023data}
Fernandes, E.; Moro, S.; and Cortez, P. 2023.
\newblock Data science, machine learning and big data in digital journalism: a survey of state-of-the-art, challenges and opportunities.
\newblock \emph{Expert Systems with Applications}, 221: 119795.

\bibitem[{Fleiss(1971)}]{fleiss1971measuring}
Fleiss, J.~L. 1971.
\newblock Measuring nominal scale agreement among many raters.
\newblock \emph{Psychological bulletin}, 76(5): 378.

\bibitem[{Freelon(2018)}]{freelon2018computational}
Freelon, D. 2018.
\newblock Computational research in the post-API age.
\newblock \emph{Political Communication}, 35(4): 665--668.

\bibitem[{Gearhart and Kang(2014)}]{gearhart2014social}
Gearhart, S.; and Kang, S. 2014.
\newblock Social media in television news: The effects of Twitter and Facebook comments on journalism.
\newblock \emph{Electronic News}, 8(4): 243--259.

\bibitem[{Google and et~al.(2023)}]{palm2}
Google; and et~al. 2023.
\newblock PaLM 2 Technical Report.
\newblock arXiv:2305.10403.

\bibitem[{Gruppi et~al.(2021)Gruppi, Adal{\i}, Salemi, and Horne}]{gruppi2021tweeting}
Gruppi, M.; Adal{\i}, S.; Salemi, M.; and Horne, B.~D. 2021.
\newblock From Tweeting About News to Creating News Around Tweets: Characterizing Tweets Embedded in News Articles.

\bibitem[{Haapanen(2020)}]{haapanen2020modelling}
Haapanen, L. 2020.
\newblock Modelling quoting in newswriting: A framework for studies on the production of news.
\newblock \emph{Journalism Practice}, 14(3): 374--394.

\bibitem[{Harry(2014)}]{harry2014journalistic}
Harry, J.~C. 2014.
\newblock Journalistic quotation: Reported speech in newspapers from a semiotic-linguistic perspective.
\newblock \emph{Journalism}, 15(8): 1041--1058.

\bibitem[{Hombaiah et~al.(2023)Hombaiah, Chen, Zhang, Bendersky, Najork, Colen, Levi, Ofitserov, and Amin}]{hombaiah2023tweembed}
Hombaiah, S.~A.; Chen, T.; Zhang, M.; Bendersky, M.; Najork, M.; Colen, M.; Levi, S.; Ofitserov, V.; and Amin, T. 2023.
\newblock Creator Context for Tweet Recommendation.
\newblock arXiv:2311.17650.

\bibitem[{Kapidzic et~al.(2022)Kapidzic, Neuberger, Frey, Stieglitz, and Mirbabaie}]{kapidzic2022news}
Kapidzic, S.; Neuberger, C.; Frey, F.; Stieglitz, S.; and Mirbabaie, M. 2022.
\newblock How News Websites Refer to Twitter: A Content Analysis of Twitter Sources in Journalism.
\newblock \emph{Journalism Studies}, 23(10): 1247--1268.

\bibitem[{Kim, Lee, and Lee(2016)}]{kim2016topical}
Kim, D.; Lee, J.-G.; and Lee, B.~S. 2016.
\newblock Topical influence modeling via topic-level interests and interactions on social curation services.
\newblock In \emph{2016 IEEE 32nd International Conference on Data Engineering (ICDE)}, 13--24. IEEE.

\bibitem[{Kwak et~al.(2010)Kwak, Lee, Park, and Moon}]{kwak2010twitter}
Kwak, H.; Lee, C.; Park, H.; and Moon, S. 2010.
\newblock What is Twitter, a social network or a news media?
\newblock In \emph{Proceedings of the 19th international conference on World wide web}, 591--600.

\bibitem[{Lawler(2022)}]{lawler2022meta}
Lawler, R. 2022.
\newblock Meta reportedly plans to shut down CrowdTangle, its tool that tracks popular social media posts.
\newblock \emph{The Verge}.
\newblock [Accessed 2023-12-04].

\bibitem[{Lin et~al.(2017)Lin, Hong, Wang, and Li}]{lin2017survey}
Lin, S.; Hong, W.; Wang, D.; and Li, T. 2017.
\newblock A survey on expert finding techniques.
\newblock \emph{Journal of Intelligent Information Systems}, 49: 255--279.

\bibitem[{Mittal et~al.(2020)Mittal, Suthar, Patil, Pranaya, Rana, and Tidke}]{mittal2020social}
Mittal, D.; Suthar, P.; Patil, M.; Pranaya, P.; Rana, D.~P.; and Tidke, B. 2020.
\newblock Social network influencer rank recommender using diverse features from topical graph.
\newblock \emph{Procedia Computer Science}, 167: 1861--1871.

\bibitem[{Mujib et~al.(2020)Mujib, Heidenreich, Murphy, Santia, Zelenkauskaite, and Williams}]{mujib2020newstweet}
Mujib, M.~I.; Heidenreich, H.~S.; Murphy, C.~J.; Santia, G.~C.; Zelenkauskaite, A.; and Williams, J.~R. 2020.
\newblock NewsTweet: a dataset of social media embedding in online journalism.
\newblock \emph{arXiv preprint arXiv:2008.02870}.

\bibitem[{Mujib, Zelenkauskaite, and Williams(2022)}]{mujib2022tweets}
Mujib, M.~I.; Zelenkauskaite, A.; and Williams, J.~R. 2022.
\newblock Which tweets deserve to be included in news stories? Chronemics of tweet embedding.
\newblock \emph{arXiv preprint arXiv:2211.09185}.

\bibitem[{Myers and Hamilton(2014)}]{myers2014social}
Myers, C.; and Hamilton, J.~F. 2014.
\newblock Social Media as Primary Source: The narrativization of twenty-first-century social movements.
\newblock \emph{Media History}, 20(4): 431--444.

\bibitem[{Oro et~al.(2017)Oro, Pizzuti, Procopio, and Ruffolo}]{oro2017detecting}
Oro, E.; Pizzuti, C.; Procopio, N.; and Ruffolo, M. 2017.
\newblock Detecting topic authoritative social media users: a multilayer network approach.
\newblock \emph{IEEE Transactions on Multimedia}, 20(5): 1195--1208.

\bibitem[{Panchendrarajan and Saxena(2023)}]{panchendrarajan2023topic}
Panchendrarajan, R.; and Saxena, A. 2023.
\newblock Topic-based influential user detection: a survey.
\newblock \emph{Applied Intelligence}, 53(5): 5998--6024.

\bibitem[{Paresh(2023)}]{paresh2023reddit}
Paresh, D. 2023.
\newblock Reddit Is Already on the Rebound.
\newblock \emph{The Verge}.
\newblock [Accessed 2023-12-04].

\bibitem[{Pei et~al.(2020)Pei, Wang, Morone, and Makse}]{pei2020influencer}
Pei, S.; Wang, J.; Morone, F.; and Makse, H.~A. 2020.
\newblock Influencer identification in dynamical complex systems.
\newblock \emph{Journal of complex networks}, 8(2): cnz029.

\bibitem[{Perriam, Birkbak, and Freeman(2020)}]{perriam2020digital}
Perriam, J.; Birkbak, A.; and Freeman, A. 2020.
\newblock Digital methods in a post-API environment.
\newblock \emph{International Journal of Social Research Methodology}, 23(3): 277--290.

\bibitem[{Porter(2023)}]{porter2023twitter}
Porter, J. 2023.
\newblock Twitter announces new API pricing, posing a challenge for small developers.
\newblock \emph{The Verge}.
\newblock [Accessed 2023-12-04].

\bibitem[{Rony, Yousuf, and Hassan(2018)}]{rony2018large}
Rony, M. M.~U.; Yousuf, M.; and Hassan, N. 2018.
\newblock A large-scale study of social media sources in news articles.
\newblock \emph{arXiv preprint arXiv:1810.13078}.

\bibitem[{Tekir et~al.(2023)Tekir, G{\"u}zel, Tenekeci, and Haman}]{tekir2023quote}
Tekir, S.; G{\"u}zel, A.; Tenekeci, S.; and Haman, B. 2023.
\newblock Quote Detection: A New Task and Dataset for NLP.
\newblock In \emph{Proceedings of the 7th Joint SIGHUM Workshop on Computational Linguistics for Cultural Heritage, Social Sciences, Humanities and Literature}, 21--27.

\bibitem[{Tromble(2021)}]{tromble2021have}
Tromble, R. 2021.
\newblock Where have all the data gone? A critical reflection on academic digital research in the post-API age.
\newblock \emph{Social Media + Society}, 7(1).

\bibitem[{Vaucher et~al.(2021)Vaucher, Spitz, Catasta, and West}]{vaucher2021quotebank}
Vaucher, T.; Spitz, A.; Catasta, M.; and West, R. 2021.
\newblock Quotebank: a corpus of quotations from a decade of news.
\newblock In \emph{Proceedings of the 14th ACM International Conference on Web Search and Data Mining}, 328--336.

\bibitem[{Vrande{\v{c}}i{\'c} and Kr{\"o}tzsch(2014)}]{vrandevcic2014wikidata}
Vrande{\v{c}}i{\'c}, D.; and Kr{\"o}tzsch, M. 2014.
\newblock Wikidata: a free collaborative knowledgebase.
\newblock \emph{Communications of the ACM}, 57(10): 78--85.

\bibitem[{Wang et~al.(2022)Wang, Wei, Schuurmans, Le, Chi, Narang, Chowdhery, and Zhou}]{wang2022self}
Wang, X.; Wei, J.; Schuurmans, D.; Le, Q.; Chi, E.; Narang, S.; Chowdhery, A.; and Zhou, D. 2022.
\newblock Self-consistency improves chain of thought reasoning in language models.
\newblock \emph{arXiv preprint arXiv:2203.11171}.

\bibitem[{Wei et~al.(2022)Wei, Wang, Schuurmans, Bosma, Xia, Chi, Le, Zhou et~al.}]{wei2022chain}
Wei, J.; Wang, X.; Schuurmans, D.; Bosma, M.; Xia, F.; Chi, E.; Le, Q.~V.; Zhou, D.; et~al. 2022.
\newblock Chain-of-thought prompting elicits reasoning in large language models.
\newblock \emph{Advances in Neural Information Processing Systems}, 35: 24824--24837.

\bibitem[{Wen et~al.(2023)Wen, Xiao, Hovy, and Hauptmann}]{wen2023towards}
Wen, H.; Xiao, Z.; Hovy, E.; and Hauptmann, A.~G. 2023.
\newblock Towards Open-Domain Twitter User Profile Inference.
\newblock In \emph{Findings of the Association for Computational Linguistics: ACL 2023}, 3172--3188.

\bibitem[{Zelizer(1989)}]{zelizer1989saying}
Zelizer, B. 1989.
\newblock `Saying' as collective practice: Quoting and differential address in the news.
\newblock \emph{Text-Interdisciplinary Journal for the Study of Discourse}, 9(4): 369--388.

\bibitem[{Zelizer(1995)}]{zelizer1995text}
Zelizer, B. 1995.
\newblock Text, talk, and journalistic quoting practices.
\newblock \emph{Communication Review (The)}, 1(1): 33--51.

\bibitem[{Zheng et~al.(2020)Zheng, Zhang, Young, and Wang}]{zheng2020demand}
Zheng, C.; Zhang, Q.; Young, S.; and Wang, W. 2020.
\newblock On-demand influencer discovery on social media.
\newblock In \emph{Proceedings of the 29th ACM international conference on information \& knowledge management}, 2337--2340.

\end{thebibliography}

\newpage

\appendix
\begin{table*}[t]\centering
  \caption{\label{tab:schema} SocialQuotes dataset schema. Topics are identified via \url{https://cloud.google.com/natural-language/docs/classifying-text}.}
  \begin{tabular}{|p{2in}|p{4.5in}|}
    \hline
    \textbf{Field} & \textbf{Description} \\ \hline\hline
    \texttt{id} & Hash uniquely identifying the (embedded post, URL) pair. \\ \hline
    \texttt{common\_crawl\_snapshot} & The Common Crawl crawl identifier\footnote{\url{https://commoncrawl.org/get-started}} from which the SM embedding was extracted. \\ \hline
    \texttt{common\_crawl\_url} & URL in Common Crawl dataset. \\ \hline
    \texttt{social\_post\_url} & The URL of the embedded SM post. \\ \hline
    \texttt{social\_user\_profile} & The URL of the social media account that created the embedded SM post. \\ \hline
    \texttt{context\_topics} & Up to three highest-probability topics identified in the page surrounding the SM embedding.\\\hline
    \texttt{role\_labels} & For a subset of quotes, each social role label identified by crowd compute raters, along with the frequency with which it was selected. \\\hline
  \end{tabular}
\end{table*}
\begin{table*}[t]
\caption{\label{tab:embedding-patterns} Per-platform HTML patterns that we used to identify SM embeddings in Common Crawl source data, and the resulting embedding counts. ``Tag Type'' refers to the type of HTML tag that encloses the embedding. ``Tag Class'' refers to the class(es) of tags with the tag type that indicate that the tag contains a SM embedding. ``Quotes'' column contains embedding count.}
\begin{tabular}{|p{0.15\linewidth}|p{0.18\linewidth}|p{0.42\linewidth}|p{0.10\linewidth}|}
\hline
Platform & Tag Type & Tag Class & Quotes\\
\hline
\hline
\multirow{2}{*}{Instagram} & \texttt{div} & \texttt{InstagramEmbedContainer} & \multirow{2}{*}{7.6M}\\
\cline{2-3}
 & \texttt{blockquote} & \texttt{instagram-media} & \\ 
 \hline
TikTok & \texttt{blockquote} & \texttt{tiktok-embed}, \texttt{tiktok\_lazy\_shortcode} & 514K \\\hline
Twitter & \texttt{blockquote} & \texttt{twitter-tweet}, \texttt{twitter-video}, \texttt{tweet-blockquote}, \texttt{twittertweet} & 24.4M \\\hline
\end{tabular}
\end{table*}
\begin{table}[t]
\centering
\caption{\label{tab:fleiss-table} \revised{Additional measurements on SocialQuotes. Fleiss' $\kappa$ computes annotator agreement across all eight roles. The final column shows the number of times, for each role $r$, that an annotator indicated both ``Other'' and $r$ for the same quote. For comparison, ``Other'' was chosen only 289 times in our annotation experiment, accounting for 1.4\% of all individual annotator responses. This means that in 210 instances ($\sim$1\%), ``Other'' was chosen by itself.}
}
\begin{tabular}{|c|c|c|}
\hline
Role  & Fleiss' $\kappa$ & ``Other'' overlap\\
\hline
\hline
AUTHORITY     & 0.226 & 0  \\ \hline
COMMENTER     & 0.319 & 13 \\ \hline
EXPERT        & 0.161 & 2  \\ \hline
INFLUENCER    & 0.095 & 28 \\ \hline
MARKETER      & 0.211 & 6  \\ \hline
SUBJECT       & 0.266 & 15 \\ \hline
SELF-PROMOTER & 0.572 & 7  \\ \hline
WITNESS       & 0.125 & 8  \\ \hline

\end{tabular}
\end{table}

\section{Appendix A: Additional dataset information}
We provide additional information about the SocialQuotes data set and the human annotation experiment.

\subsection{Pipeline details}
We provide Tables \ref{tab:schema} and \ref{tab:embedding-patterns} here to supplement our description of the pipeline in the SocialQuotes data set section.

\subsection{Data filtering}
To avoid releasing URLs that contain or point to sites with inappropriate or sensitive content, we apply two rules. First, we do not release any SocialQuotes entry that has topic classifications within the following set:
\begin{itemize}
\item /Adult
\item /Sensitive Subjects/Accidents \& Disasters
\item /Sensitive Subjects/Death \& Tragedy
\item /Sensitive Subjects/Firearms \& Weapons
\item /Sensitive Subjects/Recreational Drugs
\item /Sensitive Subjects/Self-Harm
\item /Sensitive Subjects/Violence \& Abuse
\item /Sensitive Subjects/War \& Conflict
\item /Sensitive Subjects/Other
\end{itemize}
Second, we filter any URL that contains any word from the ``List of Dirty, Naughty, Obscene or Otherwise Bad Words''\footnote{\url{https://github.com/LDNOOBW/List-of-Dirty-Naughty-Obscene-and-Otherwise-Bad-Words}}. This list contains single words and multi-word tuples. Specifically, we filter any SocialQuotes entry with a URL that satisfies either of the following two conditions:
\begin{enumerate}
\item Any single word is equivalent to any word token in the URL. The URL is word-tokenized by splitting on "-", "\_", and "." characters.
\item Any multi-word tuple joined by a "-" or a '-' character appears in the URL string.
\end{enumerate}

\subsection{Human annotation user interface}
\begin{figure*}
    \centering
    \caption{\label{fig:cc-ui-q1}The first question in the human annotation user interface. Annotators were asked if they could actually find the social quote in a current-day live rendering of the web page.}
    \includegraphics[scale=0.4]{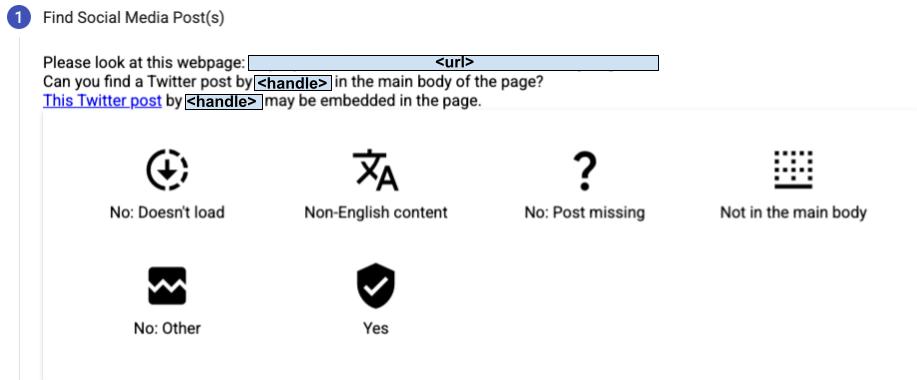}
\end{figure*}

\begin{figure*}
    \centering
    \caption{\label{fig:cc-ui-q2}The second question in the human annotation user interface. Once it was determined that the annotator could find the social quote in the page, the annotator was asked to select any number of roles they thought reflected the web author's view on the social media post. When they selected the options, short snippets appeared reminding the annotator of the role definition, which are similar to our definitions laid out in Table~\ref{tab:taxonomy}.}
    \includegraphics[scale=0.5]{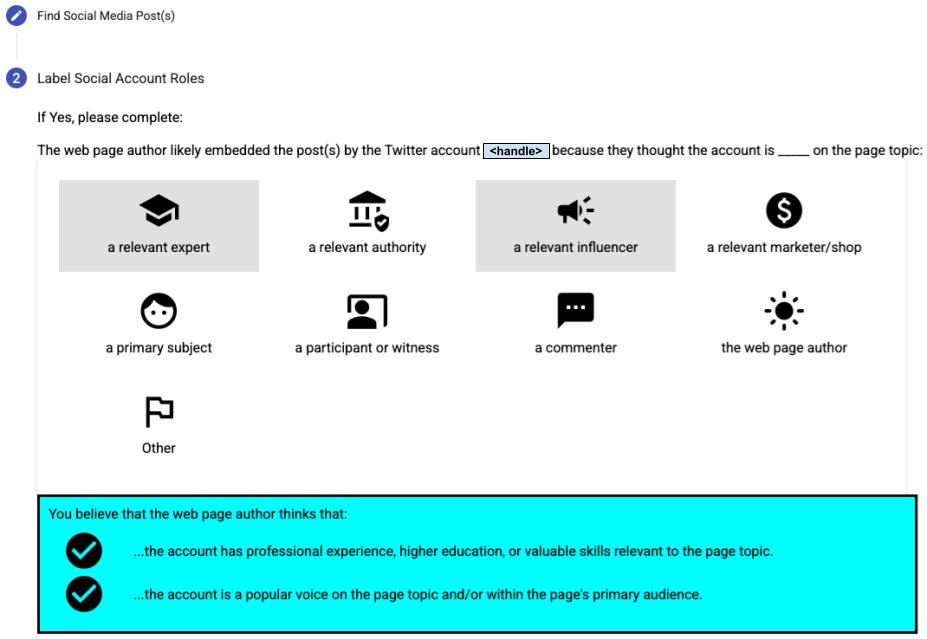}
\end{figure*}

We provide Figures \ref{fig:cc-ui-q1} and \ref{fig:cc-ui-q2} showing examples of the user interface that human annotators used to provide role annotations on social quotes.

\subsection{Role annotator agreement}
We provide a modified version of Fleiss' $\kappa$ \citep{fleiss1971measuring} to measure role annotator agreement. The Fleiss statistic has the general form $(p_o - p_e) / (1 - p_e)$, where $p_o$ is the average annotator agreement, and $p_e$ is the average annotator agreement under a random null model where each annotator chooses a label with probability proportional to the ground-truth proportion of the label. The original metric assumes that each annotator provides a annotation for every example. In our case, each annotator provides a annotation for only a subset of the examples. Taking this into account, we modify the original score by taking into account the number of annotations per example:
\begin{equation}
    p_o := \dfrac{1}{N}\displaystyle\sum_{i=1}^{N}\dfrac{1}{n_i(n_i - 1)}\displaystyle\sum_{j=1}^{k}n_{ij}(n_{ij} - 1)
\end{equation}
Above, $N$ is the number of valid annotations (see Annotation Aggregation section), $k$ is the number of classes (here $k=2$ uniformly across the eight roles), and $n_{ij}$ is the number of positive ($j=1$) or negative ($j=2$) annotations. Using this modified formula, we obtain the following Fleiss $\kappa$ metrics across roles in Table~\ref{tab:fleiss-table}.

\begin{table*}[t]
\caption{\label{tab:inference-results} Precision and recall for various models on SocialQuotes.
}
\begin{tabular}{|c|c|c|c|c|c|c|c|c|c|c|c|c|}
\hline
Model $\rightarrow$ & \multicolumn{2}{c|}{Fewshot} & \multicolumn{2}{c|}{Fewshot + SC} & \multicolumn{2}{c|}{Fewshot + P} & \multicolumn{2}{c|}{CoT} & \multicolumn{2}{c|}{CoT + SC} & \multicolumn{2}{c|}{CoT + P} \\ \hline
Role $\downarrow$ & R & P & R & P & R & P & R & P & R & P & R & P\\
\hline
AUTHORITY	 & 0.9 & 0.13 & 0.9 & 0.13 & 0.86 & 0.14 & 0.98 & 0.14 & 0.98 & 0.13 & 0.95 & 0.16 \\ \hline
COMMENTER	 & 0.66 & 0.16 & 0.68 & 0.16 & 0.6 & 0.16 & 0.96 & 0.16 & 0.98 & 0.16 & 0.92 & 0.17 \\ \hline
EXPERT	 & 0.9 & 0.1 & 0.93 & 0.1 & 0.85 & 0.11 & 0.89 & 0.11 & 0.92 & 0.1 & 0.8 & 0.12 \\ \hline
INFLUENCER	 & 0.96 & 0.3 & 0.99 & 0.3 & 0.9 & 0.3 & 0.96 & 0.3 & 0.97 & 0.3 & 0.91 & 0.31 \\ \hline
MARKETER	 & 0.53 & 0.15 & 0.54 & 0.15 & 0.48 & 0.16 & 0.79 & 0.17 & 0.85 & 0.17 & 0.71 & 0.23 \\ \hline
SUBJECT	 & 0.47 & 0.73 & 0.49 & 0.74 & 0.38 & 0.75 & 0.82 & 0.66 & 0.84 & 0.65 & 0.73 & 0.69 \\ \hline
SELF-PROMOTER	 & 0.46 & 0.72 & 0.54 & 0.76 & 0.24 & 0.79 & 0.81 & 0.52 & 0.86 & 0.5 & 0.67 & 0.78 \\ \hline
WITNESS	 & 0.49 & 0.2 & 0.49 & 0.2 & 0.46 & 0.2 & 0.85 & 0.18 & 0.89 & 0.18 & 0.75 & 0.19 \\ \hline\hline
Macro Average	 & 0.67 & 0.31 & 0.69 & 0.32 & 0.59 & 0.33 & 0.88 & 0.28 & 0.91 & 0.27 & 0.8 & 0.33 \\ \hline

\end{tabular}
\end{table*}
\begin{table*}[]
    \centering
    \caption{\label{tab:prompt-table} Role-specific prompt questions given to \llm\ in the preamble.}
    \begin{tabular}{|p{1in}|p{5in}|}
        \hline
        Role & Prompt Question \\\hline\hline
        Self-Promoter & \texttt{Your job is to determine if the embedded post was created by the same entity who created the webpage (a self-promotion).} \\\hline
        Primary-Subject& \texttt{Your job is to determine if the embedded post is from someone who is the primary entity being discussed in the webpage.} \\\hline
        Expert& \texttt{Your job is to determine if the embedded post is from someone who has recognized expertise in the main topic of the webpage.} \\\hline
        Commenter& \texttt{Your job is to determine if the embedded post is from someone who is commenting on the main topic of the webpage.} \\\hline
        Influencer& \texttt{Your job is to determine if the embedded post is from someone who is a popular voice on the main topic of the webpage.} \\\hline
        Witness-Participant& \texttt{Your job is to determine if the embedded post is from someone who witnessed or directly participated in an event discussed in the webpage.} \\\hline
        Authority& \texttt{Your job is to determine if the embedded post is from someone who is a recognized public figure or institution relevant to the webpage content.} \\\hline
        Marketer& \texttt{Your job is to determine if the webpage is marketing or advertising a product mentioned in the embedded post.} \\\hline
    \end{tabular}
\end{table*}
\begin{table*}[t]
\centering
\caption{\label{tab:domain-analysis-results} Table of co-occurrence statistics for domains vs.\ platforms and roles. Note that the top-3 roles and the (only) 3 platforms do not correspond to each other: roles and platforms are independently ranked by MI.
}
\begin{tabular}{|l||l|l|l||l|l|l|}
\hline
Domain Name & Platforms & $p$ & MI & Roles & $p$ & MI \\
\hline
\hline

sportingnews.com& Twitter & 1E+0 & 1.2 & Authority & 9E-1 & 6.0 \\
& Instagram & 2E-2 & 0.1 & Commenter & 1E+0 & 5.4 \\
& TikTok & 9E-4 & 0.1 & Expert & 9E-1 & 5.3 \\
\hline
\hline

cnet.com& Twitter & 1E+0 & 1.2 & Witness & 3E-1 & 5.6 \\
& TikTok & 2E-2 & 1.1 & Commenter & 1E+0 & 5.2 \\
& Instagram & 3E-2 & 0.2 & Influencer & 7E-1 & 4.4 \\
\hline
\hline

slate.com& Twitter & 1E+0 & 1.2 & Subject & 9E-1 & 5.9 \\
& TikTok & 1E-2 & 0.6 & Expert & 9E-1 & 5.4 \\
& Instagram & 3E-2 & 0.1 & Authority & 8E-1 & 5.3 \\
\hline
\hline

euronews.com& Twitter & 8E-1 & 1.1 & Witness & 3E-1 & 6.0 \\
& Instagram & 1E-1 & 0.8 & Authority & 8E-1 & 5.4 \\
& TikTok & 4E-3 & 0.3 & Expert & 8E-1 & 4.8 \\
\hline
\hline

allure.com& Instagram & 1E+0 & 5.1 & Marketer & 1E+0 & 30.1 \\
 & & & & Subject & 1E+0 & 6.4 \\
 & & & & Influencer & 1E+0 & 6.4 \\
\hline
\hline

vulture.com& Twitter & 8E-1 & 1.1 & Self-Prom. & 7E-1 & 6.5 \\
& Instagram & 2E-1 & 0.8 & Subject & 1E+0 & 6.2 \\
& TikTok & 4E-3 & 0.2 & Authority & 8E-1 & 5.6 \\
\hline
\hline

cnn.com& Twitter & 9E-1 & 1.1 & Subject & 9E-1 & 5.6 \\
& Instagram & 1E-1 & 0.5 & Expert & 1E+0 & 5.5 \\
& TikTok & 8E-3 & 0.5 & Authority & 8E-1 & 5.4 \\
\hline
\hline

finance.yahoo.com& Twitter & 9E-1 & 1.1 & Subject & 9E-1 & 5.6 \\
& Instagram & 1E-1 & 0.7 & Influencer & 9E-1 & 5.5 \\
 & & & & Expert & 9E-1 & 5.4 \\
\hline
\hline

techcrunch.com& Twitter & 1E+0 & 1.2 & Authority & 9E-1 & 5.9 \\
& TikTok & 9E-3 & 0.6 & Influencer & 9E-1 & 5.5 \\
& Instagram & 1E-2 & 0.0 & Commenter & 1E+0 & 5.2 \\
\hline
\hline

upworthy.com& TikTok & 7E-1 & 46.8 & Witness & 3E-1 & 6.9 \\
& Twitter & 3E-1 & 0.4 & Marketer & 2E-1 & 5.0 \\
& Instagram & 2E-2 & 0.1 & Self-Prom. & 5E-1 & 4.9 \\
\hline
\hline

\end{tabular}
\end{table*}

\begin{figure}
\centering
\includegraphics[scale=0.3]{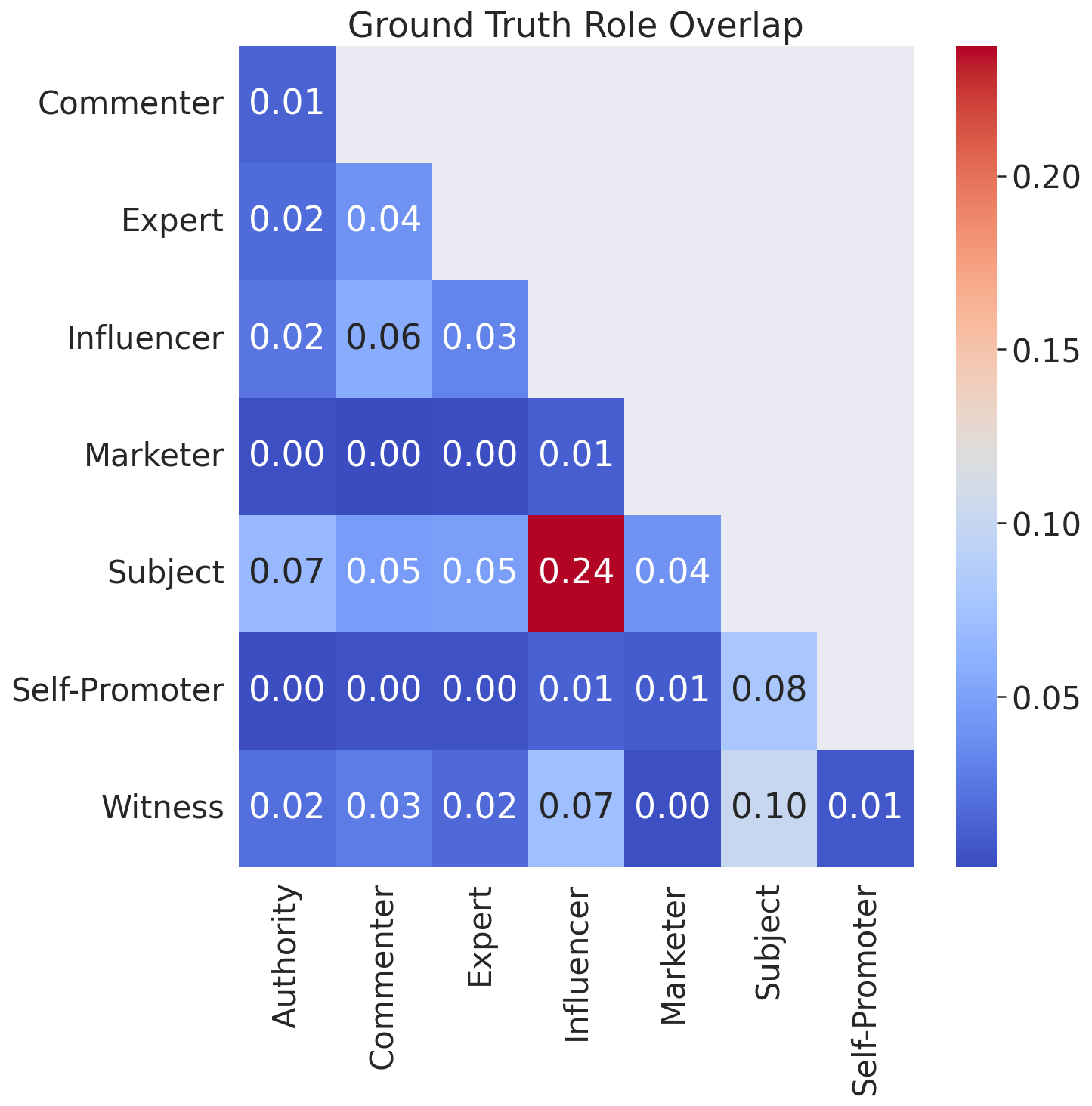}
\caption{\label{fig:role-confusion} \revised{Proportion of times each role pair co-appeared for quote with a valid ground-truth annotation.}}
\end{figure}

\begin{figure}
\includegraphics[scale=0.55]{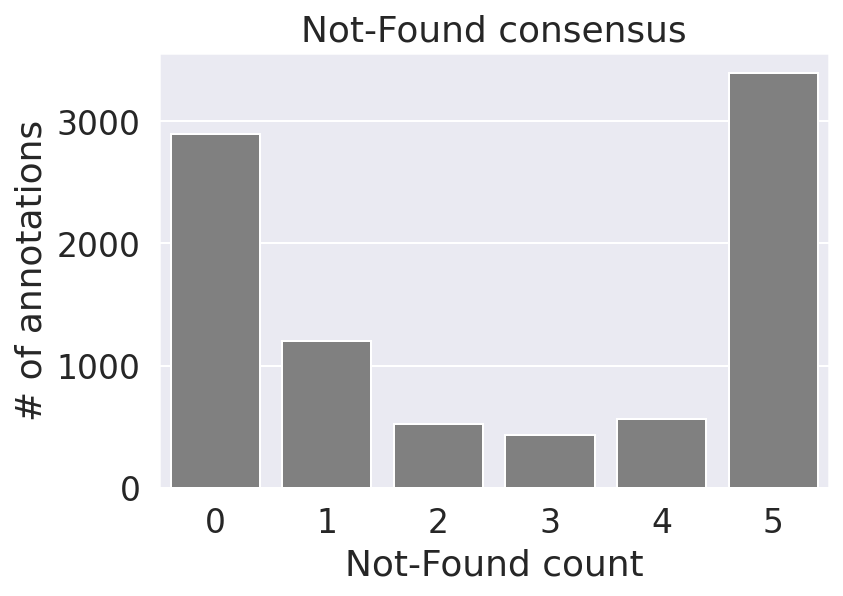}
\caption{\label{fig:not-found-counts} Number of annotations in which $k$ number of annotators could not find the quote ($k\in\{0, \ldots, 5\}$).}
\end{figure}

\subsection{\revised{Ground-truth role overlap}}
\revised{
In Figure \ref{fig:role-confusion} we show a confusion matrix of the ground-truth roles. Specifically, each matrix element is the proportion of time the two roles co-occurred across the ground-truth quote annotations. We find that the Influencer and Subject roles co-occur the most frequently, by a wide margin. Based on manual reviews of our SocialQuotes, we hypothesize that this is due to the wide number of blog/news articles that are written directly about a celebrity who is also a popular voice on a certain subject.
}
\subsection{\revised{``Other'' and ``Not Found'' annotations}}
\revised{Out of all annotations on quotes that could be located at annotation time, the ``Other'' option was selected only 289 times. Table \ref{tab:fleiss-table} counts, for each role $r$, the number of times $r$ was selected along with ``Other'' (by the same annotator that selected ``Other''). The ``Other'' annotations were not used in our experiments, however they are provided with SocialQuotes as auxiliary information.}

\revised{Figure \ref{fig:not-found-counts} shows the distribution of the number of ``Not-Found'' annotations per-quote, out of 8,281 annotated quotes not proximal to sensitive content. The majority of quotes were able to be located by all or none of the annotators. The quotes that were found only by some annotators may be (1) from websites that had locale-specific paywalls/functionality, or (2) located in parts of pages that were difficult to find for some annotators.}

\section{Appendix B: Additional experiment information}
We provide additional information about our experiments in the ``Experiments and analysis'' section.

\subsection{\revised{Prompt design}}
\revised{Below we provide the main prompt structure for PaLM-2, as well as a Chain-of-Thought example for the ``Commenter'' role}. In Table~\ref{tab:prompt-table}, we provide the complete list of prompts given to the \texttt{<role\_specific\_binary\_prompt>} field in the \llm~prompt structure presented in the ``Role inference'' section. We note that these prompts differ slightly in wording from the definitions given in Table~\ref{tab:taxonomy}, though the core meaning is the same for each role.

\subsection{Domain-based platform and role analysis}
We provide Table~\ref{tab:domain-analysis-results} as additional insight into the SocialQuotes data set. \revised{Our analysis of the results in this table can be found in the ``Social quotes analysis'' section of the main text.}

\begin{tcolorbox}[float=t,width=\linewidth,colback={lightgray},title={\llm\ prompt structure},colbacktitle=white,colback=white,coltitle=black,fontupper=\tiny]
\texttt{You are a social media analyst looking at social media posts embedded in websites. Given the following information:}\\
\vspace{0.5cm}

\texttt{URL: The URL of a webpage with an embedded social media post;}\\
\texttt{POST\_URL: The URL of the embedded social media post;}\\
\texttt{HANDLE: The social media username of the author of the embedded post;}\\
\texttt{SNIPPET: The webpage text that appears around the embedded post;}\\
\vspace{0.5cm}

\texttt{Your job is to determine if <role\_specific\_binary\_prompt>. Below are are some examples:}\\
\vspace{0.5cm}

\texttt{<fewshot\_examples>}\\
\vspace{0.5cm}

\texttt{URL: <common\_crawl\_url>}\\
\texttt{POST\_URL: <social\_post\_url>}\\
\texttt{HANDLE: <extracted\_profile\_handle>}\\
\texttt{SNIPPET: <quote\_context>}\\
\end{tcolorbox}

\begin{tcolorbox}[title={\llm\ CoT example for the ``Commenter'' role.},width=\linewidth,colback=white,coltitle=black,colbacktitle=white,fontupper=\tiny]    
\texttt{The primary focus of the snippet is the Pittsburgh Steelers football team. The post is from a fan of the team commenting on one of the players. So the embedded post is from someone commenting on the topic of the webpage.}
\end{tcolorbox}

\end{document}